\documentclass[10pt, pdflatex, sn-mathphys-num, iicol]{sn-jnl}

\usepackage{array}
\usepackage[caption=false,font=normalsize,labelfont=sf,textfont=sf]{subfig}
\usepackage{stfloats}
\usepackage{url}
\usepackage{verbatim}
\usepackage{color}
\usepackage{ragged2e}
\usepackage{etoolbox}
\usepackage{hyperref}
\usepackage{hypcap}
\usepackage{multirow}
\usepackage{multicol}
\usepackage{makecell}
\usepackage{placeins}
\usepackage{dsfont}

\usepackage{graphicx}%
\usepackage{amsmath,amssymb,amsfonts}%
\usepackage{amsthm}%
\usepackage{mathrsfs}%
\usepackage[title]{appendix}%
\usepackage{xcolor}%
\usepackage{textcomp}%
\usepackage{manyfoot}%
\usepackage{booktabs}%
\usepackage{algorithm}%
\usepackage{algorithmicx}%
\usepackage{algpseudocode}%
\usepackage{listings}%

\newcommand{\eg}[0]{\emph{e.g}.}
\newcommand{\ie}[0]{\emph{i.e}.}

\newcommand{\hj}[1]{\textcolor{black}{#1}}
\newcommand{\yl}[1]{\textcolor{black}{#1}}

\theoremstyle{thmstyleone}%
\theoremstyle{thmstyletwo}%
\theoremstyle{thmstylethree}%

\begin{document}

\title[]{RCR: Robust Crowd Reconstruction with Upright Space from a Single Large-scene Image}


\author[1]{\fnm{Jing} \sur{Huang}}\email{hj00@tju.edu.com}

\author[1]{\fnm{Hao} \sur{Wen}}\email{wenhao@tju.edu.com}

\author[1]{\fnm{Tianyi} \sur{Zhou}}\email{tianyi@tju.edu.com}

\author[2]{\fnm{Haozhe} \sur{Lin}}\email{linhz@tsinghua.edu.com}

\author[3]{\fnm{Yu-Kun} \sur{Lai}}\email{Yukun.Lai@cs.cardiff.ac.uk}

\author*[1]{\fnm{Kun} \sur{Li}}\email{lik@tju.edu.com}

\affil*[1]{\orgdiv{College of Intelligence and Computing}, \orgname{Tianjin University}, \orgaddress{\city{Tianjin}, \postcode{300350}, \country{China}}}

\affil[2]{\orgdiv{Beijing National Research Center for Information Science and Technology}, \orgname{Tsinghua University}, \orgaddress{\city{Beijing}, \postcode{10084}, \country{China}}}

\affil[3]{\orgdiv{School of Computer Science and Informatics}, \orgname{Cardiff University}, \orgaddress{\city{Cardiff}, \postcode{CF24 4AG}, \country{Unitied Kingdom}}}

\abstract{
    This paper focuses on spatially consistent hundreds of human pose and shape reconstruction \hj{from a single large-scene image with various human scales under arbitrary camera FoVs (Fields of View).}
    Due to the small and highly varying 2D human scales, depth ambiguity, and perspective distortion, no existing methods can achieve globally consistent reconstruction with correct reprojection.
    \hj{
    To address these challenges, we first propose a new concept, Human-scene Virtual Interaction Point (HVIP), to convert complex 3D human localization into 2D-pixel localization.
    We then extend it to RCR (Robust Crowd Reconstruction), which achieves globally consistent reconstruction and stable generalization on different camera FoVs without test-time optimization.}
    To perceive humans in varying pixel sizes, we propose an Iterative Ground-aware Cropping to automatically crop the image and then merge the results.
    To eliminate the influence of the camera and cropping process during the reconstruction, we introduce a canonical Upright 3D Space and the corresponding Upright 2D Space.
    To link the canonical space and the camera space, we propose the Upright Normalization, which transforms the local crop input into the Upright 2D Space, and transforms the output from the Upright 3D Space into the unified camera space.
    Besides, we contribute two benchmark datasets, \emph{LargeCrowd} and \emph{SynCrowd}, for training and evaluating crowd reconstruction in large scenes.
    Experimental results demonstrate the effectiveness of the proposed method.
    \emph{The source code and data will be publicly available for research purposes.}
}


\maketitle


\begin{figure*}
    \centering
    \includegraphics[width=\textwidth]{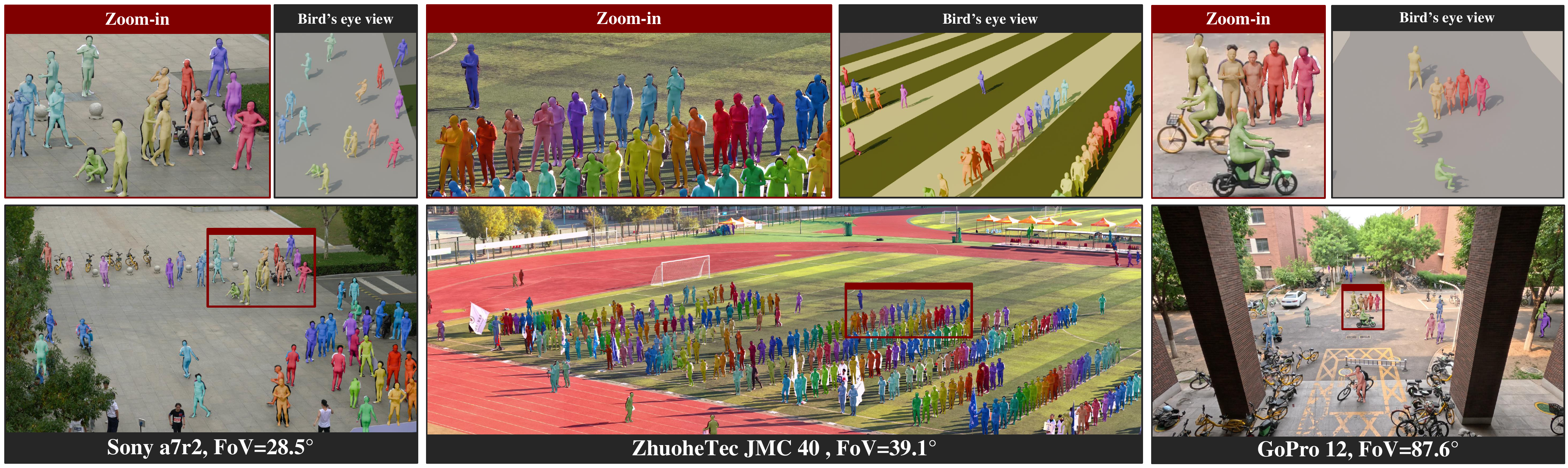}
    \caption{Given a single image under arbitrary FoVs, our method can reconstruct human poses and shapes in a unified camera space with both spatial consistency and reprojection accuracy.
    }\label{fig:teaser}
\end{figure*}

\section{Introduction}\label{sec:introduction}
3D pose, shape, and location reconstruction for hundreds of people from a single image will help model crowd behavior for simulation and security monitoring. 
In such surveillance scenes, human pixel sizes are small and variable, and the number of humans can also be large; therefore, these images can be named \emph{large-scene images}.
In this paper, we aim to reconstruct 3D poses, shapes, and locations of hundreds of people in the camera space from a single large-scene image with unknown and arbitrary camera parameters, as shown in Fig.\ {\ref{fig:teaser}}.

Existing multi-person poses and shapes reconstruction methods {\citep{BEV, ROMP, CRMH}} can estimate multiple human pose and shape parameters from a single image directly, but they can only handle small scenes where people occupy a significant portion of the image, such as a group photo.
For a large-scene image, like \emph{PANDA} {\citep{PANDA}}, with small and varying human scales, these methods will result in a reconstruction that misses a large portion of people.
On the other hand, a combination of human detection and single human reconstruction {\citep{goel2023hmr20, pymafx2023}} typically offers better adaptability to people of various scales in images.
However, this combination has a limitation in terms of reconstruction outputs being in separate 3D spaces.
Besides, adding a depth estimation branch to translate the single human reconstruction results to the camera space will lead to incorrect reprojection and relative positions of the humans.
Recently, GroupRec {\citep{GroupRec}} has made significant progress in large-scene crowd reconstruction.
GroupRec adopts the pipeline of single reconstruction after detection, and proposes a hypergraph relational reasoning network to encode the inter-human and inter-group relations.
It alleviates the reprojection issue in a regular FoV case by a test-time optimization strategy with a depth initialization {\citep{SPEC}} and additional correlations provided by the hypergraph network.

\begin{figure}[htb]
    \centering
    \includegraphics[width=0.98\linewidth]{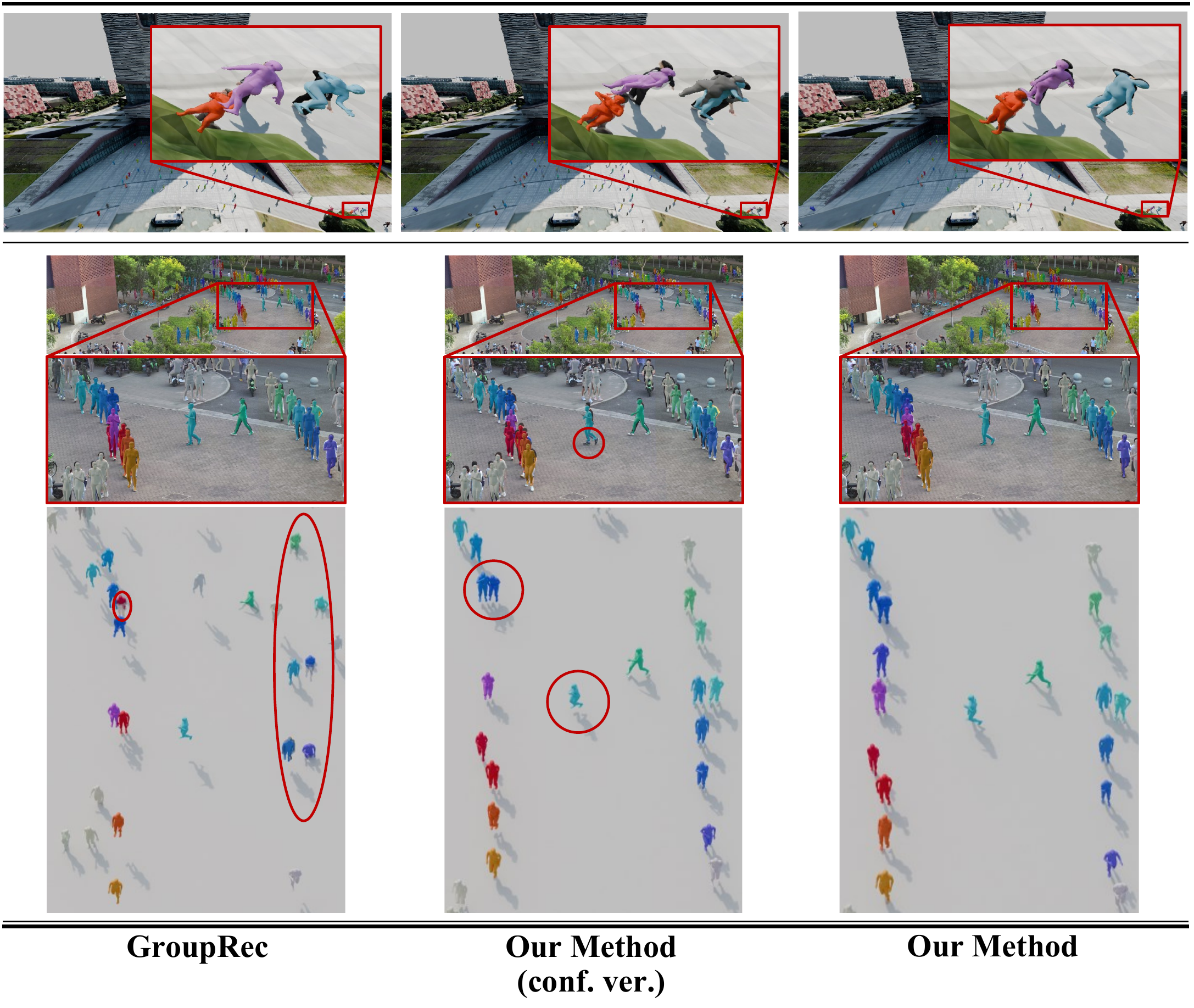}
    \caption{
        Our method achieves accurate reprojection (top line), correct relative positions, and plausibility of human-ground interaction (bottom line), while the state-of-the-art method {\citep{GroupRec}} and our conference version {\citep{Crowd3D}} do not perform well in these aspects.
    }\label{fig:sota_gap}
\end{figure}

\hj{
However, though GroupRec is enhanced by a test-time optimization with a single-plane prior and reprojection losses, it still suffers from the depth ambiguity and the perspective distortion in both regular and wide FoV cases, which leads to incorrect reprojection and relative positions of the humans, as shown in Fig.\ {\ref{fig:sota_gap}}.
}

In general, the main challenges for generalizing to arbitrary camera parameters in large-scene crowd reconstruction are:
(1) Due to the depth ambiguity from a single view, it is difficult to directly estimate absolute 3D positions and 3D poses of people in a large scene; 
(2) there are a large number of people with relatively small and highly varying 2D scales;
(3) the camera parameters and the cropping operations affect the 2D features of an actual 3D human, but learning how they affect the reconstruction results is difficult due to the lack of ground-truth 3D annotations in various camera parameters.
(4) There are no large-scene image datasets with hundreds of people to train and evaluate crowd reconstruction in large scenes.

In response to these challenges, we propose Robust Crowd Reconstruction (RCR), which achieves globally consistent crowd reconstruction in a unified camera space from a single large-scene image.
\hj{
To solve the depth ambiguity of a single image, we present a novel concept called Human-scene Virtual Interaction Point (HVIP) for effectively converting the 3D crowd spatial localization problem into a 2D HVIP estimation problem with a robust human keypoint-based camera and ground estimation approach.
To deal with a large number of people and various human scales, 
a simple idea is to crop the image into multiple local crops and then merge the results, which requires the local human pixel heights and positions as the cropping reference. 
However, the accurate local height information cannot be easily obtained by a coarse detection result, \eg{}, detection from uniformed cropped images, because the coarse detection results do not contain enough information for how to better crop the image, \ie{}, the maximum and minimum human pixel heights in a local area.
}
Therefore, we propose an Iterative Ground-aware Cropping strategy that determines the cropping sizes \hj{by the actual 3D-2D mapping} from the estimated ground plane and solves the circular dependency problem among the three processing steps: cropping, keypoints detection, and human keypoint-based camera and ground estimation.

To eliminate the influence of camera parameters and cropping operations during the single human reconstruction
we define a canonical Upright 3D Space (U3 Space) and a corresponding Upright 2D Space (U2 Space) where the ground normal is the y-axis of U3 Space, and the projection from U3 Space to U2 Space is a fixed orthographic projection.
To link the defined canonical space, we propose a Upright Normalization to transform the local crop input into the U2 Space and transform the output from the U3 Space into the unified camera space with the estimated 2D HVIP.
We also propose an HVIPNet to estimate the 2D HVIP in the U2 Space where the 2D semantic scale and the 2D orientation of a human are normalized, 
For local human pose and shape, a single human reconstruction method with weak perspective modeling estimates these parameters in the U3 Space.

We also collect and annotate \emph{LargeCrowd}, a benchmark dataset with over 100K labeled humans (2D bounding boxes, 2D keypoints, 3D ground plane, and HVIPs) in 733 gigapixel images (\(19200 \times 6480\)) of 9 scenes. To the best of our knowledge, this is the first large-scene crowd dataset that enables training and evaluation on large-scene images with hundreds of people.
Besides, we also synthesize a virtual dataset \emph{SynCrowd} at the \(9600 \times 5400\) image resolution with ground-truth SMPL {\citep{SMPL}} label by the publicly released scenes and scanned human models for quantitative testing under various camera parameters.

To summarize, our main contributions include:
\begin{itemize}
    \item We propose iterative ground-aware cropping to ensure stable human detection in various scales by the depth cues from the ground plane with a keypoints-based ground plane estimation and an iterative strategy.
    \item We propose \hj{the Human-scene Virtual Interaction Point (HVIP) concept and a keypoint-based camera and ground estimation approach} to solve the depth ambiguity problem.
    \item We propose the Upright 3D Space and Upright 2D Space with the Upright Normalization to eliminate the influence of camera parameters and cropping operations during the single-person reconstruction.
    \item We contribute \emph{LargeCrowd}, a benchmark dataset with over 100K labeled crowded people, and synthesize a virtual dataset \emph{SynCrowd} for quantitative evaluation under various camera parameters.
\end{itemize}

This work is a significant extension to our earlier conference work Crowd3D {\citep{Crowd3D}} with the following enhancements:
\begin{itemize}
    \item \textbf{Canonical Regression Space.}
    We propose the Upright 3D/2D Space and the corresponding Upright Normalization (Sec.\ {\ref{subsec:recon-via-upright}}) to eliminate the influence of camera parameters and cropping operations, while Crowd3D reconstructs humans directly in the camera space.
    \item \textbf{Automatic Cropping without Manual Settings.} 
    We propose Iterative Ground-aware Cropping to automatically determine cropping sizes by the actual 3D-2D mapping (Sec.\ {\ref{subsec:iterative-ground-aware-cropping}}), while the cropping of our conference version {\cite{Crowd3D}} determines the cropping sizes with a geometric sequence assumption and requires the maximum and minimum human pixel heights and positions as manual inputs for each image.
    \item \textbf{New Pipeline.}
    We propose a top-down pipeline with a single-person estimation network and a new HVIPNet (Sec.\ {\ref{subsubsec:hvipnet}}), while Crowd3D uses a multi-person estimation network for cropped square images. 
    \item \textbf{New Datasets.} 
    We contribute a synthesized dataset \emph{SynCrowd} with SMPL labels for quantitative evaluation under various camera parameters, since previous large-scene datasets only provide 2D keypoints annotations and images with the same camera parameters.
\end{itemize} 
These technical enhancements lead to the following improvements in effectiveness: 
(1) the reprojection performance is about \(15\%\) better than the conference version, and the bottom row of Fig.\ {\ref{fig:sota_gap}} illustrates the improvement of the reprojection and plausibility of human-ground interaction;
(2) RCR can generalize to arbitrary FoV cases without any test-time optimization, while the conference version cannot, even with a test-time optimization, as shown in the top row of Fig.\ {\ref{fig:sota_gap}};
(3) RCR can automatically process every new scene, while the conference version must manually set the cropping parameters.

\section{Related Work}\label{sec:relatedwork}
\subsection{Monocular Single Person Pose and Shape Estimation}
In most cases, human pose and shape estimation is formulated as the estimation of parameters of a parametric human model {\citep{SMPL, SMPLX2019}}.
For single-image inputs, there are optimization-based methods and regression-based methods.
Optimization-based methods are performed by fitting the SMPL model to the 2D keypoints and silhouettes {\citep{KEEPitSMPL, SMPLX2019}}.
In contrast, regression-based methods can be much faster {\citep{HMR, lhmr_lowres_xu}}, and are not limited to the SMPL parameters representation.
Some researchers propose the methods to use vertex  representation with transformers {\citep{METRO2021}}, SMPL UV representation with an image-to-image network 
{\citep{zeng2020DecoMR}}, and graph representation with graph neural networks {\citep{kolotouros2019GraphCMR}}.
The current state-of-the-art single-image methods remain the most popular approach, which is to regress the SMPL parameters using MLPs {\citep{pymafx2023}} or transformers {\citep{goel2023hmr20}} after the image feature extraction.
PyMAF-X {\citep{pymafx2023}} focuses on image alignment performance with a coarse-to-fine feedback loop, and 4D-Humans {\citep{goel2023hmr20}} exploits the general transformer structure. 
For video input, recent methods are capable of utilizing temporal information to  or output the results in camera space from videos captured by a static camera {\citep{Humor}}, 
or decouple the world space 3D pose and shape sequence and camera pose sequences from videos captured by a moving camera {\citep{gvhmr, wham, tram}} with SLAM methodology or initialization,
But this 3D localization ability of video-based methods can not be used for a single image.
In general, all these single-image methods can not output results in camera space because they use a fixed weak perspective camera model to be compatible with the various 3D and 2D datasets.

\subsection{Monocular Multi-person Pose and Shape Estimation}
A closely related task is multi-person 3D pose estimation methods {\citep{IJCV_enhanced_2021, IJCV_consensus-based_2022, IJCV_learning_2022, smap, MuCo3DHP, PAF, PANDANet}}, which can be achieved by lifting 2D joint detections to 3D space in the depth directions to ensure reprojection performance.
However, these pose-only methods sacrifice the plausibility of human pose and shape in the process of 3D lifting, so that they can not ensure important human priors, such as a natural body shape or equal-length arms and legs.
For both human pose and shape, multi-person methods  {\citep{CRMH, BEV, ROMP, BMP, 3DCrowdNet}} typically estimate the SMPL {\citep{SMPL, SMPLX2019}} pose and shape parameters to enforce human prior constraints similar to the single-person methods.
Some methods ignore the 3D locations {\citep{ROMP,3DCrowdNet}}.
ROMP {\citep{ROMP}} directly regresses multiple human SMPL parameters in a single stage.
3DCrowdNet {\citep{3DCrowdNet}} first detects 2D keypoints of all the individuals and leverages 2D poses to guide a joint-based regressor in estimating the human model parameters to address occlusions.
Some methods can achieve camera space reconstruction, but with a fixed camera focal length {\citep{CRMH, BMP, BEV}}.
CRMH {\citep{CRMH}} proposes a top-down architecture consisting of an R-CNN head and a SMPL parameter branch with additional 3D interpenetration loss and a depth ordering-aware loss to ensure the coherence of the human meshes.
BMP {\citep{BMP}} is a single-stage method with a body mesh map, which also proposes an inter-instance ordinal depth loss to constrain the depth order. 
BEV {\citep{BEV}} improves ROMP {\citep{ROMP}} by an additional bird's-eye view representation to simultaneously reason the 2D image positions and depths of the body center.
Therefore, all of the above multi-person methods can not estimate results in camera space from an entire large-scene image due to the relatively small and varying human scales compared to the image size.

\subsection{Pose and Shape Estimation from a Large-scene Image}
The above multi-person methods can not reconstruct the 3D pose and shape in a unified camera space from a single large-scene image, which includes small and varying human pixel scales and is common in surveillance scenarios.
Recently, GroupRec {\citep{GroupRec}} has made significant progress in this topic.
GroupRec proposes a hypergraph relational reasoning network to exploit the collectiveness and social interaction among crowds. 
It initializes each person's depth {\citep{SPEC}} and refines the results by a test-time optimization strategy with additional group features provided by the hypergraph network to alleviate the reprojection issue.
Our conference version Crowd3D {\citep{Crowd3D}} proposes the concept of Human-scene Virtual Interaction Point (HVIP) with a human-centric ground estimate to solve the depth ambiguity.
Specifically, it is a multi-stage method that crops the image, regresses the poses and shapes of multiple humans in the cropped image, and finally merges the results of the cropped images. Additionally, Crowd3D is also equipped with a self-supervised scene-specific optimization to refine the reprojection performance. 
However, training a network to perceive the influence of different camera FoVs and cropping positions is difficult, especially given the scarcity of datasets with diverse FoVs.
Therefore, as shown in Fig.\ {\ref{fig:sota_gap}}, the reprojection performance of Crowd3D and GroupRec is poor under arbitrary FoVs, with Crowd3D failing to ensure reasonable human-ground interactions, and GroupRec even producing incorrect relative positions or depths.

\hj{
In this paper, we propose RCR (Robust Crowd Reconstruction) with Upright Space to reconstruct multiple humans from a single large-scene image under arbitrary camera FoVs without test-time optimization.
To enable training and testing in large scenes, we contribute the \emph{LargeCrowd} dataset with human keypoints level annotations and the \emph{SynCrowd} dataset with 3D SMPL ground-truths and different camera FoVs.
}

\begin{figure*}[t]
    \centering
    \includegraphics[width=0.99\linewidth]{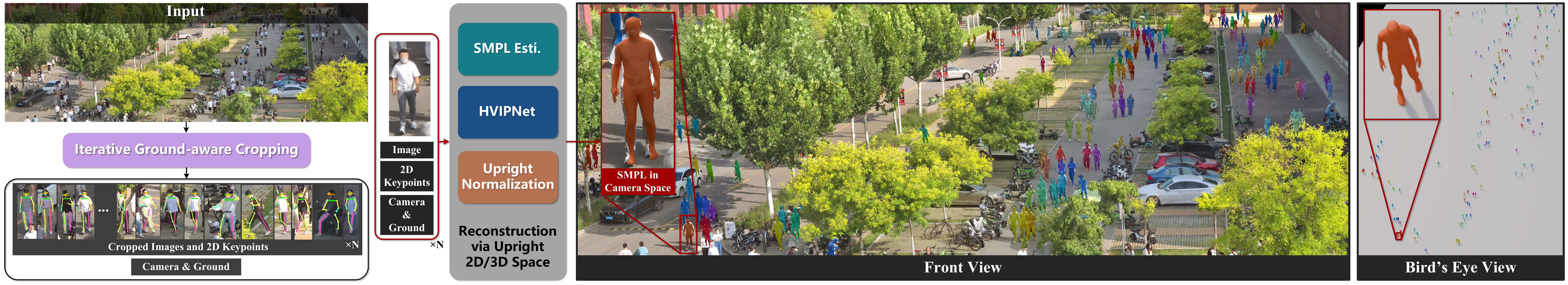}
    \caption{Overview of our method. Our method first detects all the humans and then reconstructs each person via \yl{the} upright space (details in Fig.\ {\ref{fig:upright-norm}}).
    With the help of the canonical Upright Space, though the SMPL estimation is separately conducted in the Upright 3D Space, the results can achieve both spatial consistency and reprojection accuracy in the camera space.}\label{fig:overview-rcr}
\end{figure*}

\begin{figure}[t]
    \centering
    \includegraphics[width=0.99\linewidth]{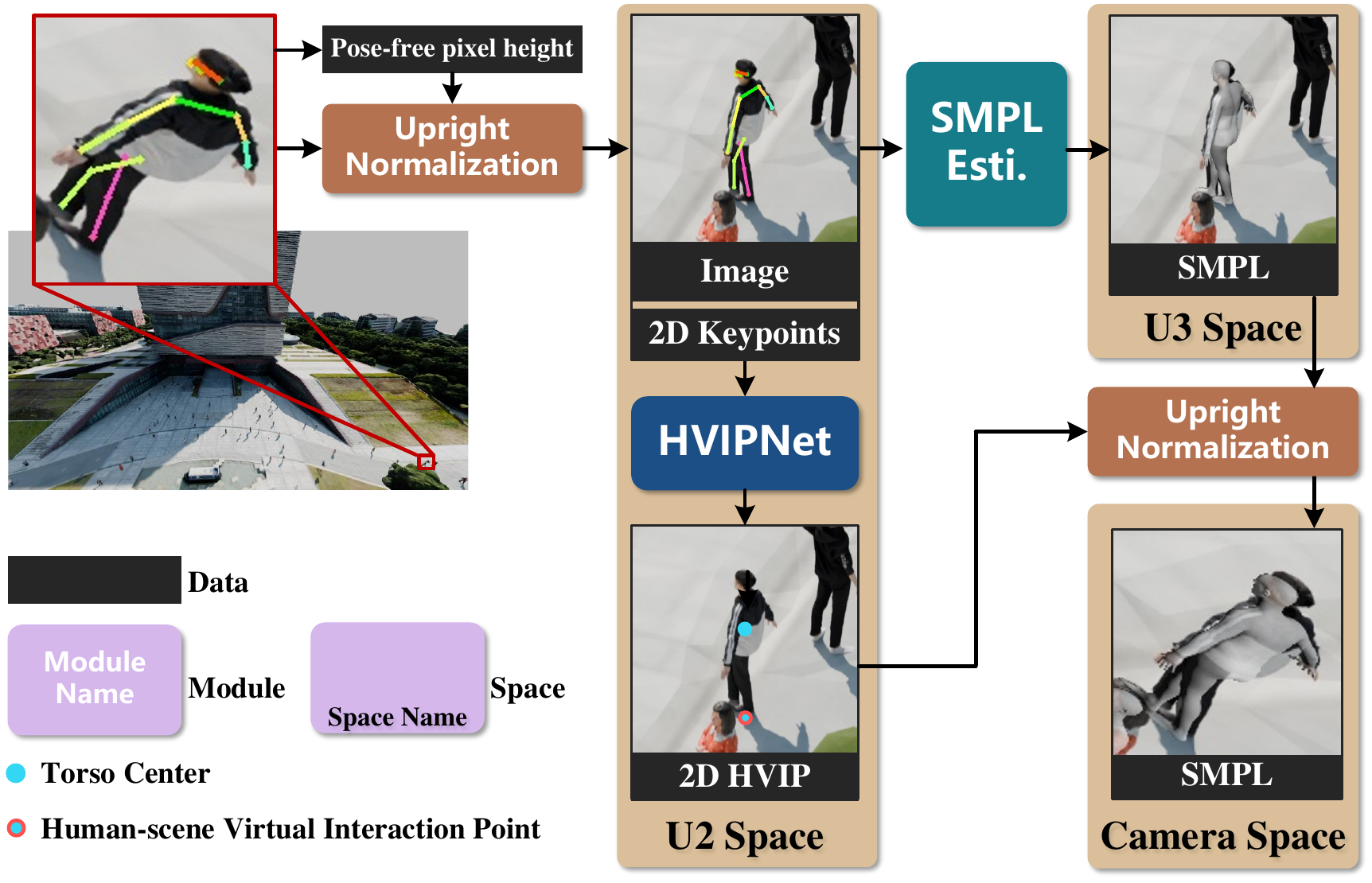}
    \caption{Reconstructing each person via Upright 2D/3D Space, Upright Normalization, and HVIPNet.}\label{fig:upright-norm}
\end{figure}

\section{Method}\label{sec:method}
\hj{
Our goal is to reconstruct 3D poses and shapes of all people from a single large-scene image as input. Our previous conference version {\citep{Crowd3D}} proposes a multi-person estimation network for cropped square images with an HVIP concept to locate each person and a simple cropping strategy that requires manually inputting the maximum and minimum human pixel heights.
}

\hj{
In this paper, we keep the HVIP concept to deal with the depth ambiguity (Sec.\ {\ref{subsec:hvipdefinition}}) and propose a new pipeline, RCR, as illustrated in Fig.\ {\ref{fig:overview-rcr}}, which first detects each individual from the cropped images and then estimates the multi-person's pose and shape parameters and the 2D HVIP of each individual.
}
To handle small and varying human scales in the image, we design an iterative ground-aware cropping 
(Sec.\ {\ref{subsec:iterative-ground-aware-cropping}}) 
to detect the keypoints of all individuals and estimate the camera and ground parameters in a loop without any manual inputs.
To eliminate the influence of the camera parameters during the reconstruction,
we reconstruct each detected human with the proposed \emph{Upright 3D Space} (U3 Space) and \emph{Upright 2D Space} (U2 Space) where the ground normal vector is parallel to the vertical axis and weak perspective projection is applicable (Sec.\ {\ref{subsec:recon-via-upright}}).
We also propose an Upright Normalization to convert the cropped image to the U2 Space with the estimated camera and ground parameters.
To locate the absolute 3D position of each individual in the camera space,
we propose an HVIPNet (Sec.\ {\ref{subsubsec:hvipnet}}) to estimate 2D HVIP (Sec.\ {\ref{subsubsec:hvipnet}}) and reconstruct the SMPL parameters (Sec.\ {\ref{subsubsec:estimate-in-upright}}) in the U3 Space.
Finally, we convert all the results back to the camera space using the proposed Upright Normalization transform so that we achieve globally consistent reconstruction in the camera space.

\subsection{Formulation}\label{subsec:formulation}
We use a pinhole camera model with a focal length \(f\ (f = f_x = f_y)\) where the principal point \((c_x, c_y)\) of the camera is the image center and assign \(K\) to be the camera intrinsic matrix.
We formulate a single and arbitrary ground plane in the camera space as:
\(N^T P_{g} + D = 0\), where \(N=(x_n, y_n, z_n)\) is the ground normal with \(\Vert N \Vert_2 = 1\), \(P_{g} \in \mathbb{R}^3\) is \yl{a} point on the ground plane and \(D \in \mathbb{R}\) constrains the distance from the camera center to the ground plane.
The tuple \(G = \{N, D\}\) is called the ground parameter.
The projection from the camera space to the image space is defined as: \(p = F_\text{proj}(K, P)\), where \(P \in \mathbb{R}^3\) is a 3D point and \(p = (u, v)\) is the 2D point in the image space. 
The reverse projection of a 2D point \(p\) from the image space to the camera space with the ground constrain is defined as: \(P_{g} = F_\text{rev-proj}(K, N, D, p_{g})\), where \(p_{g} \in \mathbb{R}^2\) is the 2D point in the image, and \(P_{g} \in \mathbb{R}^3\) is the 3D point in the camera space and on the ground plane (satisfying forward projection and \(N^T P_{g} + D = 0\)).
Our method eventually output camera space SMPL {\citep{SMPL}} parameters \(\theta \in \mathbb{R}^{72}\), shape \({\beta} \in \mathbb{R}^{10}\), scale \(S_{\text{toCam}} \in \mathbb{R}^{1}\) and \(T_{\text{toCam}} \in \mathbb{R}^{1}\), where \(\theta\) and \(\beta\) are the pose and shape parameters of SMPL, \(S_{\text{toCam}}\) and \(T_{\text{toCam}}\) are the scale and translation from the SMPL space to the camera space which should be applied after the SMPL forward function \(\mathcal{M}: (\mathbb{R}^{72}, \mathbb{R}^{10}) \rightarrow \mathbb{R}^{6890 \times 3}\).

\subsection{Human-scene Virtual Interaction Point (HVIP)}\label{subsec:hvipdefinition}
To help infer the accurate 3D locations of persons in the large-scene camera system,
we inherit the concept of \emph{Human-scene Virtual Interaction Point (HVIP)} from our conference version \citep{Crowd3D},
which is the projection point of a person's 3D torso center on the ground plane in the camera space where the 3D torso center \(P_{t} \in \mathbb{R}^3\) is defined as the center (arithmetic mean in all dimensions) of 4 primary body joints (2 shoulder joints and 2 hip joints).

\textbf{Progressive Position Transform.} 
We represent the 2D corresponding points in the image space of \(P_{v}\) and \(P_{t}\) as \(p_{v}\) and \(p_{t}\), respectively.
Since the direction of \(P_{t} - P_{v}\) is perpendicular to the ground, we have \(P_{t} = P_{v} + d \times N\), where \(d\) represents the distance from \(P_{t}\) to the ground plane.
Given a 2D torso center \(p_{t} = (u_t, v_t)\), a 2D HVIP \(p_{v} = (u_v, v_v)\), the camera and ground parameters \(K\) and \(G\), we first compute the 3D HVIP \(P_{v} = (x_v, y_v, z_v)\) by \(F_\text{rev-proj}(K, N, D, p_{v})\) since the HVIP is on the ground plane.
With the equations (1) \(P_{v} = F_\text{rev-proj}(K, N, D, p_{v})\), (2) \(P_{t} = P_{v} + d \times N\) and (3) \(p_{t} = F_\text{proj}(K, P_{t})\),
we reduce the unknowns to only one variable \(d\):
\begin{gather}
    \label{eq:solvedtorso}
    d = \frac{
        f\times y_{v} - (v_{t}-c_y)\times z_{v}
    }{(v_{t}-c_y)\times z_{n} - f \times y_{n}}.
\end{gather}
where \(f, c_{y}\) are the defined camera parameters in \(K\), and \((x_{n}, y_{n}, z_{n}) = N\) is the ground normal vector.
Therefore, the progressive position transform is a mapping:
\begin{equation}
    \label{eq:prog-pos}
    P_{t} = F_\text{prog-pos}(K, G, p_{t}, p_{v}).
\end{equation}
which means we can the 3D torso center \(P_{t}\) in the camera space by the 2D torso center \(p_{t}\) and 2D HVIP \(p_{v}\) when the camera \(K\) and ground parameters \(G\) are known. 
Since the 2D torso center can be easily extracted from the 2D keypoints and an SMPL mesh has a clear 3D torso center via the definition, we are able to locate any SMPL mesh by estimating the 2D HVIP of this person.

\subsection{Iterative Ground-aware Cropping}\label{subsec:iterative-ground-aware-cropping}
The off-the-shelf human detection models, \eg{}, MMPose {\citep{mmpose2020}}, AlphaPose {\citep{alphapose}}, and Openpose {\citep{OpenPose}}, only perform well when a person occupies a specific range of proportions in the image. 
To deal with the large-scene image with small human scales, the image cropping approach needs to ensure the detection performance.
Different from the cropping approach in our conference version {\citep{Crowd3D}}, which requires manual setting of the maximum and minimum human pixel heights, we propose an iterative ground-aware cropping strategy to automatically crop the image and merge the results.
Therefore, there are two key problems to be solved in this section:
(1) automatically deciding the size of the local cropping area, and
(2) avoiding missing and duplication after merging the results.


To address the first problem, we note that the camera and ground parameters can indicate the possible human pixel height of a local area in the image.
Camera and ground are unknown when we need to crop the original image, and estimating ground and camera may need the human 2D keypoints detection results.
This shows a circular dependency problem among the three processing steps: cropping, keypoint detection, and camera and ground estimation.
To overcome the circular dependency, we propose an iterative ground-aware cropping strategy as follows:
(1) The iterative ground-aware cropping is initialized by several attempts of uniform cropping with different scales (cropped sizes are set to 512, 1024, ..., until reaching 1/4 of the image size, thus usually only one attempt);
(2) The detection part takes the overlapped cropped images with cropping bounding boxes as input and outputs the keypoints of all people in the whole image without duplication (Sec.\ {\ref{subsubsec:detect-and-filter-duplication}});
(3) The camera and ground estimation part takes the keypoints of all individuals as input and outputs the camera and ground parameters (Sec.\ {\ref{subsubsec:camera-ground-estimation}});
(4) The ground-aware cropping part takes the camera and ground parameters as input and outputs the cropping bounding boxes.

\subsubsection{Detecting and Filtering duplication}\label{subsubsec:detect-and-filter-duplication}
This subsection takes the cropped images as input and outputs the keypoints of all individuals while avoiding duplication of the same person in the original image, since overlapping exists among the neighboring cropped images.
We run an off-the-shelf human detection toolkit {\citep{mmpose2020}} on the cropped images.
To avoid low-quality detection, we first remove detections that are truncated by the cropping boundary.
This operation is reasonable because the cropping strategy guarantees that each person will appear fully in at least one of the cropped images, and RCR does not process people truncated by the original image boundary.
Particularly, we filter out the truncated keypoints that meet both of the following conditions: (1) with at least one keypoint with a low confidence score (\(<0.2\)), (2) its corresponding bounding box area is very close to an image boundary which has a neighboring cropped image.
After this filtering step, the detection part transforms all the keypoints to the original image space.
To remove duplicated detections, we first find several groups of keypoints that are close to each other in terms of a defined 2D Pose Similarity in a KD-tree structure, where the 2D Pose Similarity is defined as the average Euclidean distance of all high-confidence (\(>0.35\)) keypoints normalized by the pixel height.
For each group, we keep the detection with the highest average confidence score to avoid duplication.
Finally, this subsection outputs the keypoints of all individuals in the original image.

\subsubsection{Keypoint-based Camera and Ground Estimation}\label{subsubsec:camera-ground-estimation}
This subsection takes the keypoints of all individuals as input and outputs the camera and ground parameters following our conference version {\citep{Crowd3D}}.
For each detected individual, we compute the midpoints of two shoulder keypoints as the top point \(p_{\text{top}} \in \mathbb{R}^2\) and the midpoints of two ankle keypoints as the bottom point \(p_{\text{bot}} \in \mathbb{R}^2\) from the detected keypoints.
We consider that (1) \(p_{\text{bot}}\) is on the ground plane, (2) the direction \(p_{\text{top}} - p_{\text{bot}}\) is close to the ground normal vector \(N\), (3) and deviation in direction follows an almost normal distribution to solve the camera \(K\) and ground parameters \(G\). Specifically, we define the reprojected top points as \(p_{\text{top}}' = F_\text{proj}(K, hN + F_\text{rev-proj}(K, N, D, p_{\text{bot}}))\), where \(h \in \mathbb{R}\) is the fixed height prior.
We then compute the loss function of one individual as:
\begin{equation}
    \begin{aligned}
        \label{eq:re-projectionLoss}
        L_{\text{g}} = & \lambda_{\text{angle}} L_{\text{cos}}( p_{\text{top}}' - p_{\text{bot}}, p_{\text{top}} - p_{\text{bot}} ) \\ 
        & + \lambda_{\text{mod}} 
        \frac {
            \left|\Vert p_{\text{top}}' - p_{\text{bot}} \Vert_2 - \Vert p_{\text{top}} - p_{\text{bot}} \Vert_2\right|
        }{\Vert p_{\text{top}} - p_{\text{bot}} \Vert_2},
\end{aligned}
\end{equation}
where \(L_{\text{cos}}\) is the cosine distance, and \(\lambda_{\text{angle}}\) and \(\lambda_{\text{mod}}\) are the weights of the corresponding loss terms. 
The two terms penalize inconsistencies between the original 2D keypoints and reprojection from angular and modular perspectives. 
Finally, we minimize the average loss of all individuals to solve the camera \(K\) and ground parameters \(G\). 
Note that this optimization can also take camera intrinsics \(K\) as an optional input and only estimate the ground plane \(G\) for better accuracy.

\begin{algorithm}[htb]
    \renewcommand{\algorithmicrequire}{\textbf{Input:}}
    \renewcommand{\algorithmicensure}{\textbf{Output:}}
    \caption{Ground-aware cropping}\label{alg:ground-aware-cropping}
    \begin{algorithmic}[1]
        \Require{}Camera intrinsic matrix \(K\), Ground plane \(G= \{ N, D \} \), \(s_{\text{minPixel}}\), maximum human height in 3D space \(H_{\text{personMax}}\)
        \Ensure{}Cropping positions \(C_\text{bbox}\)
        \State{}\(C_\text{bbox} = \emptyset \);
        \State{}\(G_\text{head} = F_\text{translate}(G, H_{\text{personMax}})\); 
            \Comment{\(F_\text{translate}\): translate the ground following the ground normal vector and get a new parallel ground plane.}
        \State{}\(d_\text{bot} = 0\); 
            \Comment{Bottom position of the current line to crop.}
        \While{True} 
            \State{}\({h}_{\text{maxPixel}} = {F}_{\text{getMaxH}}(d_\text{bot}, K, G, H_{\text{personMax}})\); 
            \If{\(h_\text{maxPixel} < s_\text{minPixel}\)} 
                \State{} \textbf{break}; \Comment{Too close to the vanishing line.}
            \EndIf{}
            \State{}\(h_\text{patch} = h_\text{maxPixel} / s_\text{ratio}\);
            \State{}\(w_\text{patch} = h_\text{patch}\);
            \State{}\(C_\text{bboxOneLine} = F_{\text{CutRow}}(d_\text{bot}, h_\text{patch}, w_\text{patch})\);
            \State{}\(C_\text{bbox} = C_\text{bbox} \cup C_\text{bboxOneLine}\);
            \State{}\(h_{\text{maxPixel}} = {F}_{\text{getMaxH}}(d_\text{bot}, K, G_\text{head}, -H_{\text{personMax}})\); 
            \State{}\(d_\text{bot} = d_\text{bot} - {h}_{\text{maxPixel}}\);
            \If{\(d_\text{bot} > H_\text{wholeImage}\)} 
                \State{} \textbf{break}; \Comment{Reach the top of the image.}
            \EndIf{}
        \EndWhile{} 
    \end{algorithmic}
\end{algorithm}

\subsubsection{Ground-aware Cropping}\label{subsubsec:ground-aware-cropping}
This subsection takes the camera \(K\) and ground parameters \(G\) as input and outputs the cropped image to ensure any person in the original image will be completely included in at least one cropped image in a proper ratio.
The key idea is to utilize the depth prior provided by the estimated camera and ground.
We set the expected human scale \(s_{\text{ratio}}\) for a cropped image 
and define the minimum human scale in pixels as \(s_{\text{minPixel}}\). 
Unlike the cropping hyperparameters in our conference version {\cite{Crowd3D}}, which need to be reset for different inputs, these values are only related to the capability of the detection toolkit. For example, when using a detection toolkit with higher precision in detecting human keypoints at very low resolutions, the value of \(s_\text{ratio}\) and \(s_{\text{minPixel}}\) may be set to a smaller value. For MMPose \citep{mmpose2020}, they are set to 0.5 and 60, respectively.
The algorithm is described in Algorithm {\ref{alg:ground-aware-cropping}} where the maximum human height in 3D space \(H_{\text{personMax}}\) can be set to 2.1 meters, function \(F_{\text{getMaxH}}\) accepts a horizontal line referred to as \(d_\text{bot}\), ground plane, camera intrinsic matrix, and the 3D length of a vector, and outputs the maximum 2D length of this vector in pixels if this vector starts from the ground plane, parallel to the ground normal and starts anywhere on this 2D horizontal line. This length can be negative which means the vector is upside down.
This algorithm returns the cropping positions \(C_\text{bbox}\) of the cropped images.

\subsubsection{Iterations}\label{subsubsec:iteration}
Since the 3 steps (Sec.\ {\ref{subsubsec:detect-and-filter-duplication}}, Sec.\ {\ref{subsubsec:camera-ground-estimation}}, and Sec.\ {\ref{subsubsec:ground-aware-cropping}) are circularly dependent, we need to iterate them until the ground parameters converge (usually 1 loop is enough) or the maximum number of iterations reaches 3.
Finally, the iterative ground-aware cropping outputs the detected keypoints of all individuals, the estimated camera \(K\), and ground parameters \(G = \{N, D\}\).

\subsection{Reconstruction via Upright 2D/3D Space}\label{subsec:recon-via-upright}
This section aims to reconstruct each individual after their keypoints are detected.
Unlike our conference version {\citep{Crowd3D}}, which directly reconstructs people in the camera space with a test-time scene-specific optimization, this paper defines a pair of canonical 2D and 3D spaces to eliminate the local and individual influence of the camera parameters during the reconstruction and achieves higher accuracy.
Specifically, we define a \emph{Upright 2D Space} (U2 Space) and the \emph{Upright 3D Space} (U3 Space) where the mapping from U3 Space to U2 Space is a fixed orthographic projection.
The U2 Space of a person is defined to satisfy the following properties: the ground normal vector is parallel to the vertical axis of the image, and the body torso center is centered horizontally in the image.
The corresponding U3 Space is defined by expanding the U2 Space in the depth direction so that the corresponding orthogonal projection is also defined.
An example of an input in the U2 Space and corresponding mesh output in the U3 Space can be seen in the Fig.\ {\ref{fig:upright-norm}}.
To link the defined U2/U3 Space with the camera space of the original image, we propose the Upright Normalization, which consists of two directions
so that the steps to reconstruct each individual can be divided into three parts: (1) converting the input image from input image space to U2 Space by the upright normalization, (2) estimating the 2D HVIP and SMPL parameters in the U3 Space, and (3) converting the results from to camera space by the upright normalization.

\subsubsection{Input Image to Upright 2D Space}
\label{subsubsec:input-to-upright}
We formulate the 2D transform from input image to the U2 Space as a homography matrix \(H_{\text{upright}} \in \mathbb{R}^{3 \times 3}\).
The homography matrix \(H_{\text{upright}}\) can be determined by 
the ground plane equation \(G\), the camera intrinsic matrix \(K\) and 
a 3D point \(P_l\) to represent where the person is located in the camera space since the depth differences between different parts of a person are tiny compared to their average depth in large scenes.
In general, we solve \(H_{\text{upright}}\) with the least squares approach after sampling specific points in the input image and computing the corresponding points in the U2 Space. 
Specifically, we first extract the 2D torso center \(p_t\) from the keypoints and get a rough 3D torso center \(P_t\) by assuming the height of this \(P_t\) from the ground to be \(1.15\) meters, where the torso center is defined in Sec.\ {\ref{subsec:hvipdefinition}}.
Although the accuracy of the \(P_t\) is rough, the following steps are enough to measure the local perspective distortion 
because the absolute depth of a person is much larger than the depth differences among different parts of this person.

We first define the basis vector in the 3D camera space as:
\( V_{\text{cam}_x} = F_\text{normalize}(N \times P_t) \), 
\( V_{\text{cam}_z} = F_\text{normalize}(P_t) \) and
\( V_{\text{cam}_y} = V_{\text{cam}_z} \times V_{\text{cam}_x} \),
where \(F_\text{normalize}\) is a function that normalizes the input vector to have a unit length. 
We then define the basis vector in the Upright 2D Space as:
\( v_{x\_\text{upright}} = [1, 0] \) and
\( v_{y\_\text{upright}} = [0, 1] \).
We sample 4 collinear points in the 3D camera space \(P_{\text{cam}_i}\) and assign the corresponding 2D points \(p_{\text{up}_i}\) in the U2 space:
\begin{align*}
    &P_{\text{cam}_1} = P_t + V_{\text{cam}_x} + V_{\text{cam}_y}, &p_{\text{up}_1} &= [1, 1], \\
    &P_{\text{cam}_2} = P_t + V_{\text{cam}_x} - V_{\text{cam}_y}, &p_{\text{up}_2} &= [1, -1], \\
    &P_{\text{cam}_3} = P_t - V_{\text{cam}_x} + V_{\text{cam}_y}, &p_{\text{up}_3} &= [-1, 1], \\
    &P_{\text{cam}_4} = P_t - V_{\text{cam}_x} - V_{\text{cam}_y}, &p_{\text{up}_4} &= [-1, -1].
\end{align*}
In a next step, we compute the 2D projected points of \(P_{\text{cam}_i}, i={1, 2, 3, 4}\) and get \(p_{\text{cam}_i}\), \ie{}, \(p_{\text{cam}_i} = F_{\text{proj}}(K, P_{\text{cam}_i})\).
Thus we can solve a homography matrix \(H_\text{warp}\) with source points \(p_{\text{cam}_i}\) and target points \(p_{\text{up}_i}\) by the least squares method. 
We set the resolution of the warped image in the U2 Space to be \(512 {\times} 512\).
For the next step, we need to scale the human subject to a specific size in the U2 Space.
To keep a semantic scale with various human poses, we define the \emph{Pose-free Pixel Height} \(h_\text{pixel}\) of a human on the image to be the sum of the trunk part, the max of the two arms and the max of the two legs of the 2D keypoints.
We then apply a scale transformation matrix to ensure that \(h_\text{pixel}\) occupies \(75\%\) of the image height and this scale transformation matrix is recorded as \(H_\text{scale} \in \mathbb{R}^{3 \times 3}\).
For the last transform step, 
we compute the translation \(H_\text{trans} \in \mathbb{R}^{3 \times 3}\)
by constraining the 2D torso center \(p_t\) to be horizontally centered at a specified distance from the top of the image.
Finally, we achieve the transform from the input image space to the Upright 2D Space
via warping, scaling, and translating: 
\(H_{\text{upright}} = H_\text{trans} \cdot H_\text{scale} \cdot H_\text{warp}\).

\subsubsection{HVIP Esitmation}\label{subsubsec:hvipnet}
According to the previous Sec.\ {\ref{subsec:hvipdefinition}}, the 3D localization problem can be converted to 2D HVIP estimation as a 2D torso center can be obtained from 2D keypoint detection by Eq. (\ref{eq:prog-pos}).
In this subsection, we estimate the 2D HVIP in the U2 Space with a proposed HVIPNet.
Different from our conference version, which has to handle various human scales in one cropped image when estimating multiple 2D HVIPs,
HVIPNet only needs to estimate the 2D HVIP of a single person in the U2 space, where the 2D ground normal is the vertical direction and a relatively constant semantic human scale ratio is kept by the proposed pose-free pixel height \(h_\text{pixel}\).
Our HVIPNet encodes an image in a fixed resolution of \(512{\times}512\) with a CNN backbone {\citep{MobileNetV3}} and flattens the feature map to a 1-D feature vector.
After feature encoding, HVIPNet applies a multi-layer perceptron (MLP) to regress a normalized scalar value \(d_\text{hvip}\) from this encoded 1-D feature vector representing the relative height from the 2D torso center to the 2D HVIP, since the ground normal is the vertical direction in the U2 space.
The number of neurons in the hidden layers of the MLP is 512, 256, 128, 64, and 32, respectively, and the activation function is LeakyReLU.
The HVIPNet is trained with the ground-truth 2D HVIP, and 2D torso center with the L1 loss on the training set of \emph{LargeCrowd} where the Upright Normalization (Sec.\ {\ref{subsubsec:input-to-upright}}) is applied by the ground truth.
The final estimated 2D HVIP in U2 Space is: \(p_{v_\text{up}} = p_t + 512 \cdot [0, d_\text{hvip}]\), which can be used to compute the 3D torso center with the estimated ground and camera parameters in Sec.\ {\ref{subsec:iterative-ground-aware-cropping}}.

\subsubsection{HVIP and SMPL Estimation in Upright 3D Space}\label{subsubsec:estimate-in-upright}
For the image transformed into the upright space for each individual, we estimate SMPL pose parameters \(\theta_{v_\text{up}}\) and shape parameters \(\beta\) with the state-of-the-art methods like {\citep{goel2023hmr20, pymafx2023}}.
These methods will also output orthographic camera parameters 
\(O = \{ {O_{s_x}}, O_{{s_y}}, O_{t_x}, O_{t_y} \} \) which describe the orthogonal projection \(u = P_x * O_{s_x} + O_{t_x}\) and \(v = P_y * O_{s_y} + O_{t_y}\), where \({O_{s_x}}, O_{{s_y}}, O_{t_x}, O_{t_y}\) are scalar values, \(P = [P_x, P_y, P_z]\) is the 3D point and \(p = [u, v]\) are the 2D projected point.
Hence, the estimated SMPL parameters \(\theta_{v_\text{up}}\) and \( \beta\) are in the U3 Space and can be projected to the U2 Space by \(O\).

\subsubsection{Upright 3D Space to Camera Space}
\label{subsubsec:upright-to-camera}
Since SMPL parameters \(\theta_{v_\text{up}}, \beta\) in U3 Space are obtained, we convert the estimated SMPL parameters back to 3D camera space by computing the additional 
scaling factor \(S_{\text{toCam}}\), 
rotation matrix \(R_{\text{toCam}}\) and 
translation vector \(T_{\text{toCam}}\)
from the previous image-to-U2 Space transforms, the local orthographic camera parameters \(O\), and the estimated 2D HVIP \(p_{v_\text{up}}\) in the 2D Upright Space.
Specifically, we first transform the upright space 2D HVIP \(p_{v_\text{up}}\) to image space \(p_v\) by the inverse matrix of \(H_{\text{upright}}\).
Secondly, according to Eq.\ ({\ref{eq:prog-pos}}), 
the 3D torso center \(P_t\) in camera space can be computed by the 2D HVIP \(p_v\), 2D torso center \(p_t\), the camera intrinsic matrix \(K\) and the ground plane \(G\).
Consequently, the translation vector \(T_{\text{toCam}}\) is the difference between \(P_t\) and the 3D torso center of the SMPL model in U3 Space.
For the rotation, \(R_{\text{toCam}}\) can be solved by 
the source basis vector \([V_{x\_\text{up}}, V_{y\_\text{up}}, V_{z\_\text{up}}] = I_{3}\) and 
the target basis vector \([V_{\text{cam}_x}, V_{\text{cam}_y}, V_{\text{cam}_z}]\).
The scaling factor can be determined by equating the local orthogonal projection with the global perspective projection of a planar object.
Since \(O_{s_x} \yl{\equiv} O_{s_y}\) in state-of-the-art methods, the scaling factor \(S_{\text{toCam}}\) can be 1-D and solved by the following equation:
\begin{equation}
\begin{aligned}
    O_{s_y} = \Vert & F_\text{project}(K, P_t + 0.5 \cdot N) \\
    - & F_\text{project}(K, P_t - 0.5 \cdot N) \Vert_{2},
\end{aligned}
\end{equation} 
where \(F_\text{project} \in (\mathbb{R}^{3 \times 3} \times \mathbb{R}^{3}) \rightarrow \mathbb{R}^{2}\) is a function that projects a 3D point to the 2D image space by camera \(K\).
We incorporate the additional rotation into the SMPL pose parameters and update the camera translation accordingly to simplify the outputs.
Therefore, the final outputs for each individual are the 3D pose \(\theta_{cam}\), shape \({\beta}\), scale \(S_{\text{toCam}}\) and translation \(T_{\text{toCam}}\).

\section{Experiments}\label{sec:experiments}

\subsection{Large-scene Datasets}\label{subsec:datasets}

To train and evaluate crowd reconstruction in a large scene, we contribute \emph{LargeCrowd}, 
which is a benchmark dataset with over 100K labeled humans in 733 gigapixel images (\(19200\times6480\)) of 9 different scenes (5 scenes for training and 4 scenes for testing).
The images are extracted at a minimum interval of \(3\) seconds from gigapixel streams, which are captured by a ZoheTec JMC315 array camera.
We annotate the bounding boxes, 2D poses, and 2D HVIPs of all the visible people in the images, 
with the maximum error less than 5 pixels for \( 95\% \) labels. 
We measure 3D landmarks in a world coordinate system and label the corresponding 2D points to solve the camera extrinsic matrix for each scene.

\subsubsection{SynCrowd}\label{subsubsec:SynCrowd}
To evaluate the accuracy of 3D joints and generalization on different camera FOVs, we create a synthetic dataset named \emph{SynCrowd}.
We collect 26 scenes from GigaMVS {\citep{GigaMVS}} and BlendedMVS {\citep{BlendedMVS}} datasets
and use 2447 scanned human models from Thuman {\citep{thumanfunction4d}} and 2K2K {\citep{2K2K}} datasets.
For each scene, we set several camera poses and a planar area for random human placement.
The human scans are placed in the area with random rotations and positions where the lowest points contact the scene and minimum distances between any two humans are constrained.
We render 68 images at \(9600{\times}5400 \) image resolution with camera FoV varying from 30 to 120 degrees with Blender {\citep{SoftwareBlender}} and its physically based rendering engine Cycles.
\emph{SynCrowd} is for testing only.

\subsection{Evaluation Metrics}
\subsubsection{Metrics for Mutually Exclusive and Collectively Exhaustive.}
The first and indispensable step of evaluating a multi-human reconstruction method is to match the predicted 3D human poses and shapes to the ground truth.
However, it is not appropriate to count the matched pairs only because it would encourage the method to predict fewer people with higher confidence. 
The best results should be mutually exclusive and collectively exhaustive (MECE), \ie{}, there is no missing or redundant prediction.
Therefore, we follow BEV {\citep{BEV}}, which adopts the F1 score to show the MECE performance, punishes other metrics by the F1 score, and reports both the matched value and the value punished by the F1 score.
The F1 score is the harmonic mean of precision and recall. In our evaluation, we exclude the cases occluded by the scene and truncated by the image edge when computing precision and recall. The details can be found in the supplementary material.

\subsubsection{Metrics for Sptail Consistency.}
To evaluate the spatial consistency of the reconstructed crowd, we use the percentage of correct ordinal depth (PCOD) {\citep{smap}} to evaluate the ordinal depth relations between all pairs of individuals.
We further propose pair-wise percentage distance similarity (PPDS) 
and the Procrustes-aligned version (PA-PPDS) to evaluate the relative position distribution of the crowd. 
The pair-wise percentage distance similarity (PPDS) is defined as
\begin{gather}
    \text{PPDS}=
    \frac{\sum_{k=1}^{n-1} \sum_{i=k+1}^{n} 1- \min{(d_{ik}, 1)}}{C_n^{2}}, \\
    d_{ik}=\left | \frac{ \Vert E_k - E_i \Vert -\Vert G_k - G_i \Vert }{\Vert G_k - G_i \Vert}\right |,
\end{gather}
where \(n\) is the number of people in the image, 
and \(E_i\) and \(G_i\) represent the estimated and ground-truth body torso \yl{centers} of the \(i\)-th person respectively.
To eliminate the influence of the scale,
we also define the Procrustes-aligned pair-wise percentage distance similarity (PA-PPDS), which aligns the reconstructed crowd and the ground truth by Procrustes alignment.
In the quantitative tables of this paper, the metrics are presented in both ``match'' and ``norm'' versions.
``Match'' means the average metrics of each individual in all the images, and ``norm'' refers to normalized, meaning the weighted average metrics that are punished by the F1 score of each image.
\hj{Since metrics PCOD, PPDS, and PA-PPDS are scores (a higher score is better)},
the way of punishment is \(\text{Score}_{\text{punished}} = \text{Score} \times \text{F1}\).

\subsubsection{Metrics for Human Poses and Shapes.}
Due to the \yl{unavoidable} depth-ambiguity and the unknown camera intrinsic parameters,
the widely-used mean per joint position error (MPJPE) exceeds the meaningful range (\(>2m\)) for all the methods.
Therefore, we adopt T-MPJPE and PA-MPJPE to evaluate the accuracy of the 3D joints, where the T-MPJPE is the MPJPE after aligning the root joints only by translation, and the PA-MPJPE is the MPJPE after Procrustes' Analysis.
We also use the object keypoint similarity (OKS) {\citep{coco}} to evaluate the reprojected 2D joint accuracy.
These metrics are also given in both ``match'' and ``norm'' versions, the same as the spatial consistency metrics.
\hj{Since OKS, T-MPJPE, and PA-MPJPE are errors (a lower error is better)}, the way of punishment is \(\text{Error}_{\text{punished}} = \text{Error} / \text{F1}\).

\begin{table*}[htbp]
    \small
    \centering
    \def\up{\( \uparrow \)}
    \def\down{\( \downarrow \)}
    \newcommand{\s}[1]{\textbf{#1}}
    \caption{Quantitative comparison on the \emph{LargeCrowd} dataset. The first 4 metrics report in both ``norm'' and ``match'' values.}
    \resizebox{\textwidth}{!}{%
    \begin{tabular}{cccccccc}
        \toprule
        \textbf{Metric} & \makecell[c]{PyMAF-large\\\citep{pymafx2023}} & \makecell[c]{4DHumans-large\\\citep{goel2023hmr20}} & \makecell[c]{BEV-large\\\citep{BEV}} & \makecell[c]{GroupRec\\\citep{GroupRec}} & \makecell[c]{Our Method\\(conf. ver.)} & \makecell[c]{Our method} \\
        \midrule
        PPDS\up{} & 70.35 / 73.62 & 73.11 / 75.90 & 66.34 / 73.35 & 71.58 / 76.27 & 83.81 / 87.27 & \s{88.92} / \s{92.01} \\
        PA-PPDS\up{} & 72.73 / 76.12 & 74.47 / 77.31 & 68.46 / 75.73 & 73.34 / 78.15 & 91.80 / 95.60 & \s{92.76} / \s{95.98} \\
        PCOD\up{} & 83.79 / 87.72 & 84.63 / 87.87 & 79.67 / 88.21 & 83.41 / 88.90 & 94.46 / 98.38 & \s{94.92} / \s{98.22} \\
        OKS\up{} & 70.53 / 73.81 & 80.54 / 83.63 & 67.08 / 74.28 & 78.40 / 83.55 & 73.55 / 76.59 & \s{82.49} / \s{85.36} \\
        Recall\up{} & 96.06\(\%\) & 96.1\(\%\) & \s{97.5\(\%\)} & 94.6\(\%\) & 94.6\(\%\) & 96.5\(\%\) \\
        Precision\up{} & 95.1\(\%\) & 96.6\(\%\) & 84.3\(\%\) & 93.2\(\%\) & \s{97.5\(\%\)} & 96.9\(\%\) \\
        F1\up{} & 0.955 & 0.963 & 0.903 & 0.938 & 0.960 & \s{0.966} \\
        \bottomrule
    \end{tabular}%
    }
    \label{tab:comp_on_LargeCrowd_est_cam}
\end{table*}

\begin{table*}[htbp]
    \def\up{\( \uparrow \)}
    \def\down{\( \downarrow \)}
    \newcommand{\s}[1]{\textbf{#1}}
    \centering
    \small
    \caption{Quantitative comparison on \emph{SytheticCrowd} dataset. The first 6 metrics report in both ``norm'' and ``match'' values.}
    \resizebox{\textwidth}{!}{%
    \begin{tabular}{ccccccc}
        \toprule
        \textbf{Metric} & \makecell[c]{PyMAF-large\\\citep{pymafx2023}} & \makecell[c]{4DHumans-large\\\citep{goel2023hmr20}} & \makecell[c]{BEV-large\\\citep{BEV}} & \makecell[c]{GroupRec\\\citep{GroupRec}} & \makecell[c]{Our Method\\(conf. ver.)} & \makecell[c]{Our Method} \\

        \midrule
        PPDS\up{} & 84.25 / 85.93 & 85.26 / 86.61 & 71.04 / 75.04 & 81.60 / 85.58 & 84.03 / 87.14 & \s{89.92} / \s{91.29} \\
        PA-PPDS\up{} & 86.91 / 88.65 & 87.76 / 89.13 & 72.71 / 76.80 & 83.84 / 87.92 & 90.60 / 93.94 & \s{93.58} / \s{95.00} \\
        PCOD\up{} & 91.48 / 93.30 & 92.10 / 93.54 & 81.99 / 86.64 & 88.65 / 92.99 & 94.04 / 97.49 & \s{96.28} / \s{97.74} \\
        OKS\up{} & 79.57 / 81.16 & 83.47 / 84.78 & 70.37 / 74.27 & 84.79 / 88.89 & 72.06 / 74.56 & \s{85.30} / \s{86.61} \\
        T-MPJPE\down{} & 0.150 / 0.147 & 0.131 / 0.129 & 0.154 / 0.145 & 0.121 / 0.115 & 0.142 / 0.135 & \s{0.112} / \s{0.110} \\
        PA-MPJPE\down{} & 0.0942 / 0.0923 & 0.0850 / 0.0836 & 0.0798 / \s{0.0753} & 0.0858 / 0.0814 & 0.0951 / 0.0910 & \s{0.0794} / 0.0782 \\
        Recall\up{} & 98.6\(\%\) & 98.5\(\%\) & \s{99.4\(\%\)} & 97.9\(\%\) & 96.7\(\%\) & 98.4\(\%\) \\
        Precision\up{} & 97.6\(\%\) & 98.4\(\%\) & 90.6\(\%\) & 93.1\(\%\) & 96.7\(\%\) & \s{98.6\(\%\)} \\
        F1\up{} & 0.980 & \s{0.985} & 0.947 & 0.953 & 0.965 & \s{0.985} \\
        \bottomrule
    \end{tabular}%
    }
    \label{tab:comp_on_SynCrowd_est_cam}
\end{table*}
\begin{figure*}[ht]
    \centering
    \includegraphics[width=0.8\linewidth]{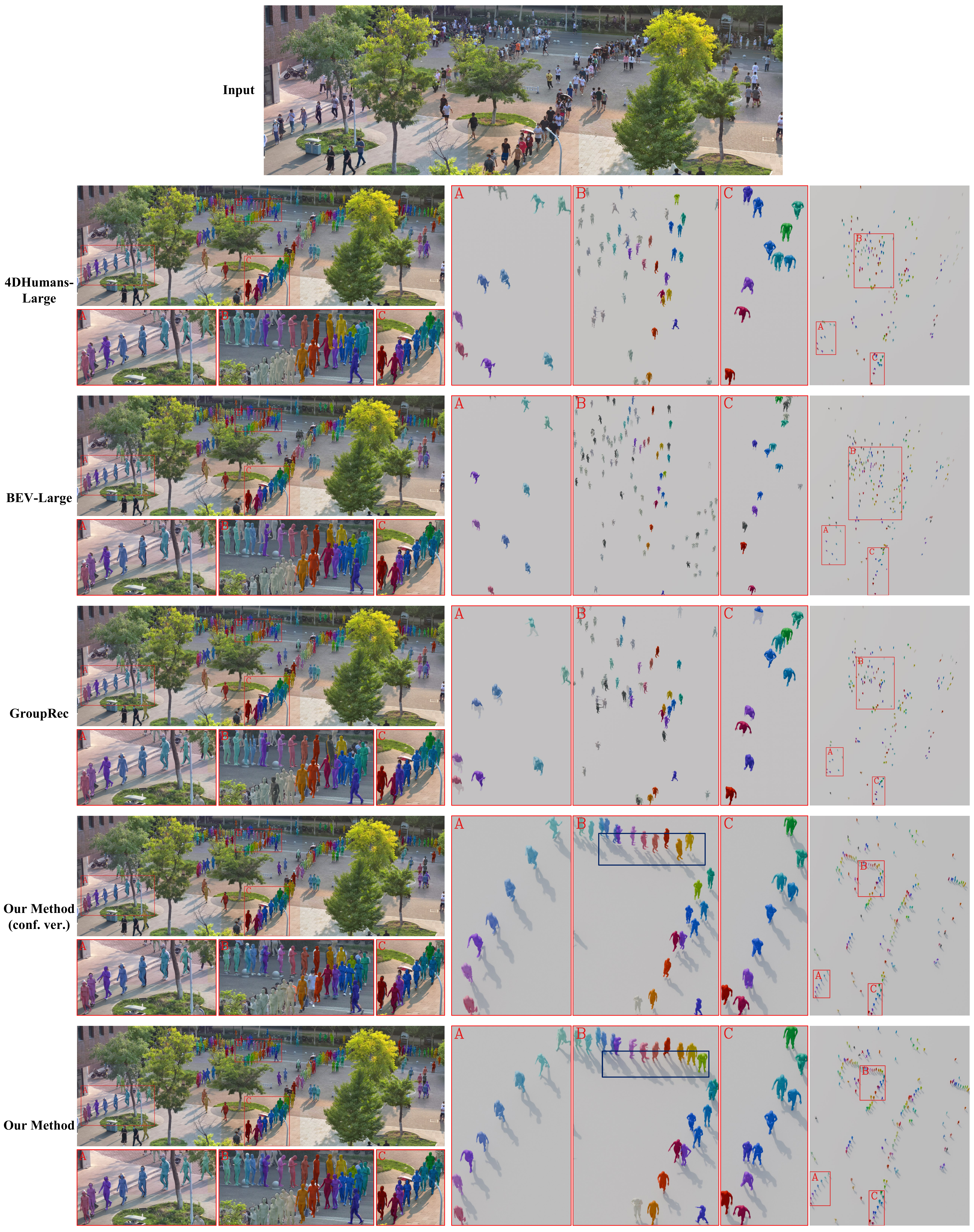}
    \caption{
        Qualitative comparison on the \emph{LargeCrowd} dataset. 
        The color of a reconstructed human corresponds to the matched ground truth, while unmatched individuals are shown in gray. 
        In the zoomed-in images A, B, and C, we use color saturation to differentiate whether the reconstructed results are fully within the cropped sub-images, \ie{}, the lower saturation indicates that the result is not within the area. 
        Shadows are ignored while rendering some compared methods because the results exhibit significant offsets from a unified ground plane.
    }\label{fig:comp_on_largecrowd}
\end{figure*}

\begin{figure*}[ht]
    \centering
    \includegraphics[width=0.9\linewidth]{comp_on_sc_quality_main_text.pdf}
    \caption{Qualitative comparison on the \emph{SynCrowd} dataset.}\label{fig:comp_on_SynCrowd_quality}
\end{figure*}

\begin{figure*}[ht]
    \centering
    \includegraphics[width=0.75\linewidth]{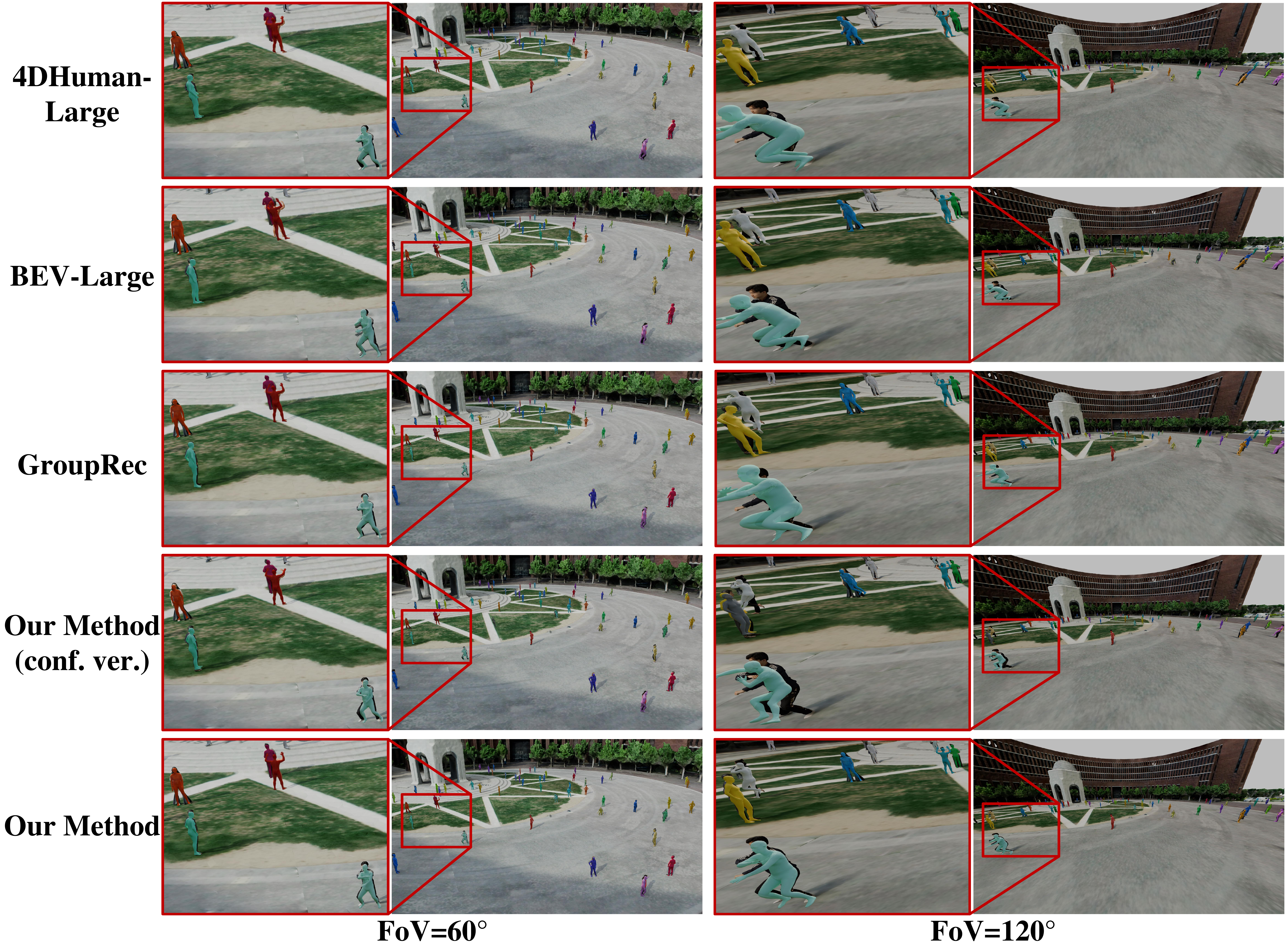}
    \caption{Qualitative comparisons on the \emph{SynCrowd} dataset under different FoVs.}\label{fig:comp_on_SynCrowd_fov}
\end{figure*}

\subsection{Comparison}\label{subsec:comparison}
\subsubsection{Compared Method}\label{subsec:largemethod}
Currently, only our methods and GroupRec {\citep{GroupRec}} can handle large-scene images with small human sizes compared to the image size.
To enable other state-of-the-art methods to process large-scene images,
we equip the state-of-the-art multi-human reconstruction method {\citep{BEV}} with our iterative ground-aware cropping in Sec.\ {\ref{subsec:iterative-ground-aware-cropping}} and the state-of-the-art single human reconstruction method 4DHumans {\citep{goel2023hmr20}} and PyMAF {\citep{pymafx2023}} with the same detection results as our method.
We use the solve-PnP algorithm to compute the \yl{position} for each individual, which is also used in the comparison experiments of BEV {\citep{BEV}}.
At last, we employ the same strategy as our method to remove redundant people for BEV's outputs.
\hj{
The names of enhanced methods are marked with ``-large'' suffix, and detailed modifications can be found in the supplementary material.
}

\subsubsection{Evaluation}\label{subsec:evaluation}
For a fair comparison, 
our method and GroupRec {\citep{GroupRec}} are trained on \emph{LargeCrowd} and BEV{\citep{BEV}} is finetuned on the cropped dataset of \emph{LargeCrowd} (cropped by Sec.\ {\ref{subsec:iterative-ground-aware-cropping}}).
Note that we finally chose the original weights of BEV for the best performance. A possible reason for the performance drop is that only the 2D keypoints are available during the fine-tuning.
Our method uses the pre-trained 4DHumans {\citep{goel2023hmr20}} as the single human reconstruction method in the upright space.
The compared 4DHumans-large uses the same pre-trained model.
None of the methods are trained on \emph{SynCrowd}, and we only use it for testing.
Because the compared methods must take the focal length as the input, these methods take the estimated camera intrinsics by our method as the input to compare fairly.
Table {\ref{tab:comp_on_SynCrowd_est_cam}} and Table {\ref{tab:comp_on_LargeCrowd_est_cam}} present the quantitative results on the \emph{SynCrowd} dataset and the \emph{LargeCrowd} dataset, respectively.
The tables show that our method is optimal in terms of both unpenalized metrics (``match'') and normalized metrics (``norm'').
The advantage in PPDS, PA-PPDS, and PCOD shows that RCR achieves better spatial consistency, including more accurate physical distances and better relative arrangements.
Fig.\ {\ref{fig:comp_on_SynCrowd_quality}} and Fig.\ {\ref{fig:comp_on_largecrowd}} show the qualitative comparison on the these datasets.
The visualization of \yl{PyMAF} {\citep{pymafx2023}} is ignored because the space is limited and it has already been compared by 4DHumans {\citep{goel2023hmr20}}.
It can be seen that only RCR and our conference version can provide accurate
relative positioning among individuals (bird's-eye view). Besides, RCR is much better in terms of reprojection, 3D poses, and interaction with the ground (highlighted by the dark blue bound boxes) against the conference version.
Besides, benefiting from the canonical upright space, our method can infer more reasonable human poses and shapes with better reprojection under different FoVs, as demonstrated in Fig.\ {\ref{fig:comp_on_SynCrowd_fov}}.

\begin{table*}[htp]
    \small
    \centering
    \def\up{\( \uparrow \)}
    \def\down{\( \downarrow \)}
    \newcommand{\s}[1]{\textbf{#1}}
    \caption{
        \yl{Ablation} studies on the \emph{LargeCrowd}. Note that ``Our Method with PyMAF'' refers to our RCR pipeline with PyMAF~\citep{pymafx2023} while ``PyMAF-large'' in the previous tables is a modified large-scene pipeline, \ie{}, Pymaf enhanced by a human detection stage. The first 4 metrics report in both ``norm'' and ``match'' values.
    }
    \begin{tabular*}{\textwidth}{@{\extracolsep{\fill}}cccc}
        \toprule
        \textbf{Metric} & \makecell[c]{Our Method w/o \\Upright Normalization} & \makecell[c]{Our Method with\\PyMAF~\citep{pymafx2023}} & \makecell[c]{Our Method with\\4DHumans~\citep{goel2023hmr20}} \\
        \midrule
        PPDS\up{} & 87.68 / 91.07 & 88.01 / 91.72 & \s{88.92} / \s{92.01} \\
        PA-PPDS\up{} & 91.37 / 94.89 & 91.76 / 95.63 & \s{92.76} / \s{95.98} \\
        PCOD\up{} & 94.16 / 97.79 & 94.09 / 98.07 & \s{94.92} / \s{98.22} \\
        OKS\up{} & 77.08 / 80.07 & 74.72 / 77.87 & \s{82.49} / \s{85.36} \\
        Recall\up{} & \s{96.7\(\%\)} & 96.4\(\%\) & 96.5\(\%\) \\
        Precision\up{} & 95.9\(\%\) & 95.6\(\%\) & \s{96.9\(\%\)} \\
        F1\up{} & 0.963 & 0.959 & \s{0.966} \\
        \bottomrule
    \end{tabular*}
\label{tab:ablation_on_LargeCrowd}
\end{table*}

\begin{table*}[htp]
    \small
    \def\up{\( \uparrow \)}
    \def\down{\( \downarrow \)}
    \newcommand{\s}[1]{\textbf{#1}}
    
    \centering
    \caption{\yl{Ablation} studies on the \emph{SynCrowd} dataset. The first 6 metrics report in both ``norm'' and ``match'' values.}
    \begin{tabular*}{\textwidth}{@{\extracolsep{\fill}}cccc}
        \toprule
        \textbf{Metric} & \makecell[c]{Our Method w/o \\Upright Normalization} & \makecell[c]{Our Method with\\PyMAF~\citep{pymafx2023}} & \makecell[c]{Our Method with\\4DHumans~\citep{goel2023hmr20}} \\
        \midrule
        PPDS\up{} & 88.86 / 90.66 & 89.45 / 91.08 & \s{89.92} / \s{91.29} \\
        PA-PPDS\up{} & 92.18 / 94.03 & 93.08 / 94.78 & \s{93.58} / \s{95.00} \\
        PCOD\up{} & 95.56 / 97.47 & 95.90 / 97.66 & \s{96.28} / \s{97.74} \\
        OKS\up{} & 77.15 / 78.71 & 80.35 / 81.81 & \s{85.30} / \s{86.61} \\
        T-MPJPE\down{} & 0.132 / 0.129 & 0.138 / 0.135 & \s{0.112} / \s{0.110} \\
        PA-MPJPE\down{} & 0.0832 / 0.0815 & 0.0932 / 0.0915 & \s{0.0794} / \s{0.0782} \\
        Recall\up{} & \s{98.7\(\%\)} & 98.5\(\%\) & 98.4\(\%\) \\
        Precision\up{} & 97.5\(\%\) & 98.0\(\%\) & \s{98.6\(\%\)} \\
        F1\up{} & 0.980 & 0.982 & \s{0.985} \\
        \bottomrule
    \end{tabular*}
\label{tab:ablation_on_SynCrowd_est_cam}
\end{table*}

\begin{table*}[htp]
    \def\up{\( \uparrow \)}
    \def\down{\( \downarrow \)}
    \newcommand{\s}[1]{\textbf{#1}}
    \centering
    \caption{Ablation Studies for the efficacy of the Iterative Ground-aware Cropping.}
    \begin{tabular}{lcccccc}
    \toprule
    \multicolumn{1}{c}{\multirow{2}[0]{*}{Method}} & 
    \multicolumn{3}{c}{On \emph{LargeCrowd}} & 
    \multicolumn{3}{c}{On \emph{SynCrowd}} \\
        & \multicolumn{1}{l}{Recall \up{}} & 
        \multicolumn{1}{l}{Precision \up{}} & 
        \multicolumn{1}{l}{F1 \up{}} & 
        \multicolumn{1}{l}{Recall \up{}} & 
        \multicolumn{1}{l}{Precision \up{}} & 
        \multicolumn{1}{l}{F1 \up{}} \\
    \midrule
    Uniform Cropping & 0.953 & 0.961 &  0.957 & 0.869 & 0.965 & 0.906 \\
    Iterative Ground-aware Cropping & \textbf{0.965} & \textbf{0.969} & \textbf{0.966} & \textbf{0.965} & \textbf{0.969} & \textbf{0.966} \\
    \bottomrule
    \end{tabular}%
    \label{tab:abla_cropping}%
\end{table*}%

\subsection{Ablation Study}\label{subsec:ablation}
\subsubsection{Efficacy of the Upright Spaces}
In RCR, the defined Upright 3D/2D Spaces and the proposed Upright Normalization are designed to eliminate the effect of the camera parameters and the cropping operation on the global reconstruction.
Although Fig.\ {\ref{fig:comp_on_SynCrowd_fov}} has demonstrated the effectiveness of the upright normalization.
We also design a variant, ``RCR w/o Upright Normalization'', which directly regresses SMPL from the cropped images with the same cropping strategy.
To minimize the difference, ``RCR w/o Upright Normalization'' is provided with the same HVIPs as the full version of RCR for 3D localization.
Table {\ref{tab:ablation_on_SynCrowd_est_cam}} reports that the full version of RCR outperforms the ablation version regarding all metrics. 
Because the same 2D HVIPs are used, the ablation version has a close performance in terms of PPDS, PA-PPDS, and PCOD.{\@} The full version has considerable advantages in terms of OKS, T-MPJPE, and PA-MPJPE, which indicates that the defined Upright 3D/2D Space is essential for human pose and shape reconstruction in large scenes.

\subsubsection{Efficacy of the Iterative Ground-aware Cropping}\label{subsubsec:abla-crop}
\hj{In RCR, we design the iterative ground-aware cropping to detect the keypoints of all individuals to handle small and varying human scales in a large-scene image. 
To verify its efficacy, we make a uniform cropping baseline, which selects cropping windows roughly twice the pixel height of the largest person in each dataset, with a \(50\%\) overlap. For \emph{LargeCrowd} and \emph{SynCrowd}, the maximum pixel heights are about \(1200\) and \(1500\) so that the cropping window sizes are \(2400\) and \(3000\), respectively.
We evaluate the detection performance of the iterative ground-aware cropping and the uniform cropping baseline with the same detection strategy in Table {\ref{tab:abla_cropping}}. The results demonstrate that the iterative ground-aware cropping consistently outperforms the baseline across all metrics, without requiring any manual settings.}

\subsubsection{Compatibility with SOTA Reconstruction Methods.}\label{subsubsec:recon_engine}
RCR can be compatible with any state-of-the-art \yl{(SOTA)} single-human reconstruction methods trained in weak perspective modeling, such as 4DHumans {\citep{goel2023hmr20}} and \yl{PyMAF} {\citep{pymafx2023}}.
They are noted as RCR with 4DHumans and RCR with \yl{PyMAF} respectively, and their results are illustrated in the Table {\ref{tab:ablation_on_LargeCrowd}} and Table {\ref{tab:ablation_on_SynCrowd_est_cam}}.
The tables show that our method maintains stable positioning performance (PPDS, PA-PPDS, and PCOD) with different single-human reconstruction methods.
The original performance of \yl{PyMAF} can be referred to as PyMAF-large in Table {\ref{tab:comp_on_LargeCrowd_est_cam}} and Table {\ref{tab:comp_on_SynCrowd_est_cam}}.
In terms of joint accuracy (OKS, T-MPJPE, and PA-MPJPE), the accuracy of our method is positively correlated with the accuracy of the single human reconstruction method.
Additionally, as the performance of the chosen method improves (the OKS of 4DHumans-large is better than that of \yl{PyMAF}-large), the performance of our method is also enhanced (RCR with 4DHumans is better than RCR with \yl{PyMAF} according to Table {\ref{tab:ablation_on_SynCrowd_est_cam}, \ref{tab:ablation_on_LargeCrowd}}).

\begin{figure}[ht]
    \centering
    \includegraphics[width=0.99\linewidth]{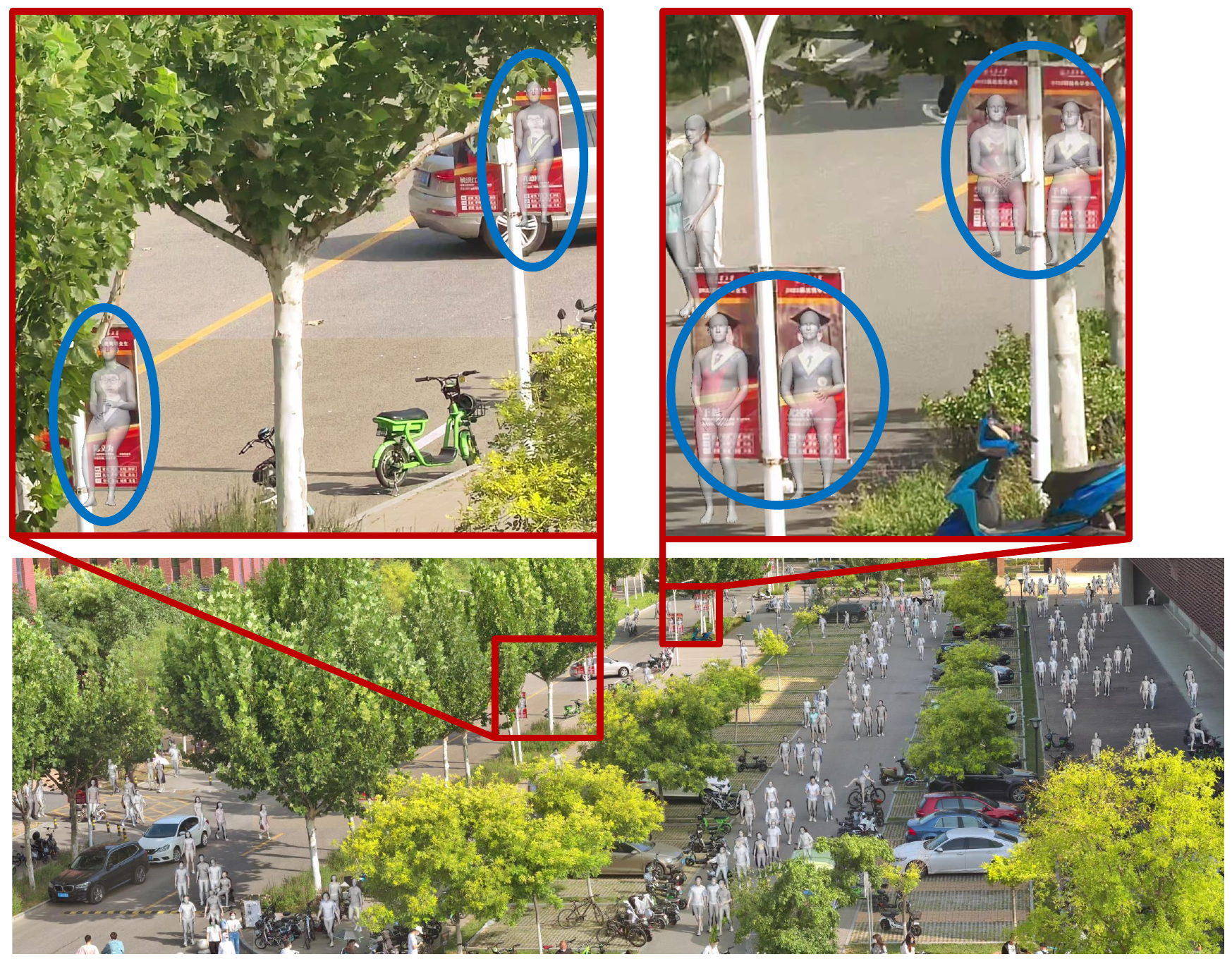}
    \caption{Failure cases of redundant detection. It is challenging to distinguish whether these human figures are fake or real.}\label{fig:fail_redundant}
\end{figure}

\subsection{Discussions}
We focus on the single-image crowd reconstruction from surveillance-like scenes.
Since we adopt the Human-scene Virtual Interaction Point to solve the depth ambiguity, our method relies on the existence of a ground plane, which is common in a surveillance scene.
For complex ground structures, the localization process of HVIP requires further designs to handle complex scenes like multiple floors and stairs by fusing the information of additional provided or estimated scene models and camera poses.
Moreover, our method is based on a standard perspective camera model, which may not be directly applicable to certain types of cameras, such as fisheye or catadioptric cameras.
However, once the camera model is identified and calibrated, upright normalization could potentially be extended to accommodate these camera types.

\hj{
\textbf{Failure Cases}. In rare cases, the Iterative Ground-aware Cropping may detect wrong objects as humans, \eg{},
Although the proposed Iterative Ground-aware Cropping can ensure a proper human scale for the human detection method while avoiding additional missing and redundant detection caused by the cropping and merging, our method can not remove some detection errors in the reconstruction stage, as shown in Fig.\ {\ref{fig:fail_redundant}}.
Our method can remove extremely big or small detections by the depth cue, but it still relies on the detection results to some extent.
}

\section{Conclusion}\label{sec:conclusion}
To reconstruct multiple human poses and shapes in the camera space from a single large-scene image, we first propose the HVIP concept to effectively solve the depth ambiguity with a human keypoint-based camera and ground estimation approach.
We then propose RCR (Robust Crowd Reconstruction) to reconstruct multiple human poses and shapes in a unified camera space from a single large-scene image. 
The proposed Upright 3D/2D Space and Upright Normalization eliminate the influence of the camera parameters and the cropping process during the reconstruction and ensure the reprojection performance under various FoVs without any test-time optimization or manually setting cropping parameters.
We also contribute the \emph{LargeCrowd} and \emph{SynCrowd} datasets to help train and evaluate crowd reconstruction in large scenes under various camera parameters.
Experimental results demonstrate that our method can achieve globally consistent crowd reconstruction in large scenes under various camera FoVs.

\bibliography{egbib}
\end{document}